\documentclass{article}
\usepackage{spconf,amsmath,amssymb,graphicx,hyperref}
\usepackage{booktabs,tabularx,array,multirow}
\usepackage{algorithm}
\usepackage{algpseudocode}
\usepackage{placeins}
\usepackage{url}
\usepackage{eso-pic}
\hypersetup{hidelinks}
\algrenewcommand\algorithmicindent{0.85em}

\title{IMPROVING IMAGE-BASED NUTRITION ESTIMATION THROUGH MULTIMODAL FOOD-ITEM VERIFICATION AND RECOVERY}
\name{Jingbo Yue, Bruce Coburn, Jinge Ma, Jui-Feng Chi, Fengqing Zhu\thanks{Corresponding author: Fengqing Zhu (zhu0@purdue.edu).}}
\address{Elmore Family School of Electrical and Computer Engineering, Purdue University\\
West Lafayette, IN, USA \quad \{yue53, coburn6, ma859, chi96, zhu0\}@purdue.edu}

\begin{document}

\AddToShipoutPictureFG*{%
  \AtPageLowerLeft{%
    \put(\LenToUnit{0.75in},\LenToUnit{0.32in}){%
      \parbox[b]{\dimexpr\paperwidth-1.5in\relax}{%
        \centering
        \fontsize{7}{8}\selectfont
        This work has been submitted to the IEEE for possible publication.\\
        Copyright may be transferred without notice, after which this version may no longer be accessible.
      }%
    }%
  }%
}

\ninept
\maketitle

\begin{abstract}
Single-image nutrition estimation can fail silently when visible foods are missed. Even when a food is correctly identified, its proposed region may not support portion estimation. We propose a framework that uses multimodal large language models (MLLMs) to inventory visible foods and separately verify food identity and whether each proposed 2D region supports portion estimation. One whole-image review uses these verification results to identify unresolved gaps and omitted foods, triggering at most one targeted recovery pass. Recovered regions are re-verified without access to the recovery prompt, then reconciled into a final item set for nutrition estimation. The framework requires no task-specific fine-tuning. Matched evaluation on common valid-output samples shows that item-level grounding improves mass accuracy across all tested settings and energy accuracy relative to an adapted retrieval baseline, with item-identity precision and recall also improving, while post-recovery visual coverage is assessed separately at inference time without ground-truth annotations.

\end{abstract}

\section{Introduction}
Meal images rarely present foods as isolated, well-separated objects. Touching foods, small side items, liquids with weak visual cues for portion estimation, and mixed dishes with components that are not independently portionable make single-view dietary assessment particularly challenging. Multimodal large language models (MLLMs) offer a natural interface for jointly interpreting food appearance and contextual cues. However, a single holistic pass can miss fine or contextually difficult visual evidence; guided visual search has shown benefits in visually crowded scenes \cite{wu2024vstar}. In dietary assessment, missed visible foods can remain unaccounted for in downstream portion and nutrition estimation.

Recent systems improve nutrition estimation by structuring intermediate reasoning or grounding meal descriptions to nutritional databases \cite{khlaisamniang2025decomposing,yan2025dietai24}. However, for single-view 2D dietary images, downstream nutrition estimation requires two different types of evidence for each visible food: reliable semantic identity for database grounding and usable image-space evidence for portion reasoning. These requirements are not equivalent: a region may support food identity while being too partial for portion estimation, whereas a well-localized region may still have ambiguous food identity. Our framework explicitly audits both before downstream nutrition estimation.

Our central design principle is that each segmentation proposal used to localize an inventory item is \emph{fallible evidence, not item truth}. Each proposal is independently verified along these two evidence axes, and two explicit support checks identify initial food inventory items lacking either type of evidence. One whole-image review, conditioned on the verified evidence state, then addresses these gaps while also identifying visible foods omitted from the initial food inventory. Unsupported initial food hypotheses can be suppressed or left unresolved, while visually supported gaps receive at most one targeted recovery attempt. Recovery comprises one round of targeted re-segmentation and independent re-verification without recovery phrases or target provenance. The resulting evidence is reconciled deterministically into the final item set used for downstream food-code and portion estimation.

Our contributions are twofold. First, we introduce a single-round item-grounded verification and recovery framework that audits semantic and geometric evidence before downstream estimation, without task-specific fine-tuning. Second, we establish a multi-level protocol spanning pre-reconciliation evidence, post-reconciliation identity quality, and matched downstream nutrition accuracy to test whether grounding improves nutrition estimates, with post-recovery visual coverage assessed separately without ground-truth labels. 
\begin{figure*}[!t]
    \centering
    \includegraphics[
        trim=0in 0in 0in 0.09in,
        clip,
        width=\textwidth,
        height=3.15in,
        keepaspectratio
    ]{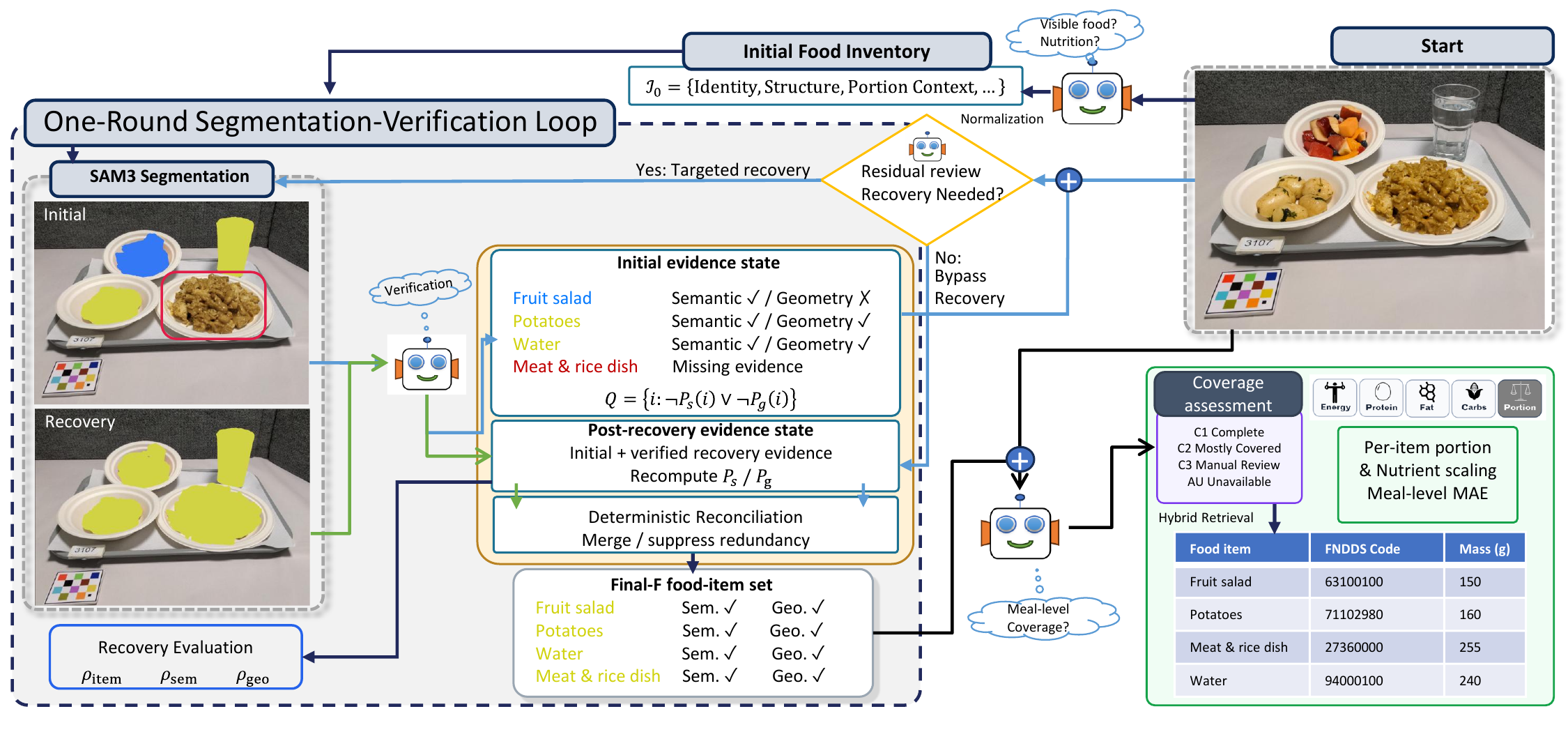}
    \caption{One-round item-grounded recovery. Verified initial and recovery evidence are re-audited and reconciled into $F$. Coverage assessment and nutrition estimation follow. The example shows fruit-salad geometry repair and meat-and-rice evidence recovery.}
    \label{fig:pipeline}
\end{figure*}
\section{Related Work}

\textbf{Dietary MLLMs and nutrition grounding.}
Recent dietary MLLM systems increasingly structure intermediate reasoning rather than directly estimating nutrition from an image. A nutritionist-inspired two-step approach separates food analysis from numerical estimation \cite{khlaisamniang2025decomposing}, while DietAI24 grounds meal descriptions to the Food and Nutrient Database for Dietary Studies (FNDDS) through retrieval before portion and nutrient inference \cite{yan2025dietai24}. DietDelta further performs food-item-level reasoning from paired before-and-after images to estimate consumed amounts \cite{vinod2026dietdelta}. Unlike these systems, we explicitly verify and, when necessary, recover the visual food-item representation before downstream grounding and portion inference. 

\textbf{Active visual search and verification.}
Guided visual search can recover evidence missed by holistic perception \cite{wu2024vstar}, while DeepEyes, VLM-R$^3$, and Woodpecker further explore active or post-hoc visual verification \cite{zheng2026deepeyes,jiang2025vlmr3,yin2024woodpecker}. Dietary estimation differs because recovered evidence must support both food identity for database grounding and geometry for portion reasoning. We therefore use one whole-image re-inspection driven by explicit evidence gaps and hide recovery provenance before re-verification.

\textbf{Food segmentation and portion evidence.}
FoodSeg103 provides large-scale food segmentation supervision \cite{wu2021foodseg}, while earlier dietary systems used recognition feedback to refine segmentation hypotheses \cite{zhu2011segassist,zhu2015multiplehypotheses,wang2018context}. We instead use open-vocabulary MLLM/SAM evidence to recover missing foods and reconcile food identity with geometry for portion estimation. Accurate food isolation is central to volume-oriented pipelines such as VolTex \cite{almughrabi2025voltex}, while monocular approaches use 3D scaling or point-cloud reconstruction \cite{vinod2024objectscaling,ma2025mfp3d}. Accordingly, segmentation serves as evidence for item-level verification and recovery within the downstream nutrition pipeline.

\section{Methodology}

\subsection{Item proposals and evidence audit}
\label{sec:itemaudit}

Fig.~\ref{fig:pipeline} shows our framework. Given a meal image $I$, an MLLM produces a structured initial food inventory. Each \emph{nutrition item} must be semantically meaningful, independently localizable, and independently portionable; inseparable mixed foods remain atomic. Deterministic normalization preserves uncertainty and portion cues, assigns stable IDs, and flags parent/component overlap risks.

\noindent\textbf{Proposal generation and verification}:
For each normalized item $t_i$, SAM~3 \cite{carion2026sam3} receives a deterministic short phrase at confidence threshold $0.50$; thereafter, scores serve only as provenance. Each proposal is shown to an MLLM verifier as a gray-background masked target with a $15\%$-padded context crop. The verifier independently assesses food identity, inventory assignment, segmentation quality, target scope, and portion geometry; the SAM prompt does not determine assignment. Geometry denotes 2D image-space region evidence, not explicit depth or volume reconstruction.

\noindent\textbf{Item-level evidence audit}:
Proposal-level verification results are summarized into two item-level indicators:
\begin{align}
P_s(i) &= \mathbf{1}[\text{verified semantic support exists}],\\
P_g(i) &= \mathbf{1}[\text{grouped geometry is provisionally ready}],
\end{align}
where $\mathbf{1}[\cdot]$ is the indicator function. $P_s(i)=1$ if at least one successfully verified proposal is assigned to item $i$; this indicates semantic support, not final semantic readiness. 
 
To determine $P_g(i)$, we retain successfully verified, in-scope proposals with correct or partial segmentation and a resolved geometry type. At least one must have correct segmentation and portion-usable geometry. Any other proposal with mask $A$ must overlap at least one such supporting mask $B$ with IoU$\geq0.80$ or
IoM$=|A\cap B|/\min(|A|,|B|)\geq0.90$. For each $A$, let $U_A$ be the union of supporting masks $B$ satisfying either overlap threshold; we additionally require $|A\cap U_A|/|A|\geq0.90$. The known-item gap set is
\begin{equation}
\boxed{
Q=\{i:\neg P_s(i)\lor\neg P_g(i)\}.
}
\label{eq:queue}
\end{equation}
Since $Q$ is restricted to initial inventory items, entirely omitted foods require nomination by the whole-image residual review.

\subsection{Residual review, recovery and reconciliation}
Every meal receives one whole-image residual review conditioned on the verified evidence state. It receives the unmodified image, initial food inventory, $Q$, and accepted-proposal summaries, but no nutrition-database information, ground truth, or proposal-source identity. For each known gap it may \emph{recover}, \emph{suppress unsupported}, or \emph{abstain unresolved}, and it may nominate omitted visible foods.

Residual review retains the original image: subtracting fallible
proposal masks could irreversibly remove neighboring or overlapping
visual evidence. Recovery lowers the SAM threshold from $0.50$
to $0.10$; verification and reconciliation criteria remain unchanged. Within one recovery round, each target is segmented once and re-verified without recovery provenance.

After source-blind re-verification, verified initial and recovery
proposals are combined and re-audited. Deterministic reconciliation
merges evidence sharing the same identity, suppresses parent/component redundancy, and leaves ambiguous geometry unresolved. Mask count does not determine serving count.

The resulting representation is
\begin{equation}
\boxed{
F=\{f_i\}_{i=1}^{K},\qquad
f_i=\{id_i,s_i,G_i,r_i^{\rm sem},r_i^{\rm geo},\Pi_i\},
}
\label{eq:finalF}
\end{equation}
where $id_i$, $s_i$, $G_i$, and $\Pi_i$ denote ID, semantics, union geometry, and provenance; $r_i^{\rm sem}$, $r_i^{\rm geo}$ indicate semantic and geometric readiness.

After $F$ is frozen, a ground-truth-free whole-image MLLM assigns meal-level Coverage: C1 (\emph{visually complete}), C2 (\emph{mostly covered}), or C3 (\emph{needs manual review}); AU denotes unavailable assessment. C1 does not imply full semantic and geometric readiness for every item. Coverage is read-only, uses no numerical threshold, and cannot modify $F$ or trigger further recovery. C1+C2 forms the broader downstream visual cohort, while C1 denotes the visually complete subset used for the primary matched nutrition comparison.

\subsection{Food-code grounding and nutrition}

A deterministic adapter maps each eligible Final-$F$ item to a per-item query while preserving $id_i$ and provenance $\Pi_i$. Each item is independently grounded to FNDDS using the hybrid retrieval and constrained food-code selection in Sec.~\ref{sec:implementation}. The selected code is immutable; legal portions are obtained by exact-code lookup with no semantic retrieval or cross-code fallback and presented with item-local geometry and preserved portion cues.

The portion model selects a legal portion and a positive multiplier
or abstains. Abstention is terminal: no default weight is imputed,
and missing grams for a required Final-$F$ item mark the meal
incomplete.  Let $F_{\rm eff}\subseteq F$ contain items with a valid
code, legal portion, and positive multiplier. For $i\in F_{\rm eff}$, $\hat g_i=w(p_i)m_i$, $\hat y_{i,q}=\hat g_i v_{c_i,q}/100$, and $\hat Y_{\mathrm{meal},q}=\sum_{i\in F_{\rm eff}}\hat y_{i,q}$,
where $c_i$, $p_i$, $w(p_i)$, and $m_i$ denote the code, portion,
portion weight, and multiplier; $v_{c_i,q}$ is nutrient $q$ per
100~g. Meal mass is $\sum_{i\in F_{\rm eff}}\hat g_i$.

\section{Experimental Setup}
\label{sec:experiments}

\subsection{Datasets, models, and implementation} 
\label{sec:implementation}
The main experiments use 1,658 Nutrition5k (N5k) images obtained with the public DietAI24 preprocessing code \cite{thames2021nutrition5k,yan2025dietai24}, retaining the full preprocessed pool rather than the 1,000-image evaluation subset reported in DietAI24, and 790 ACETADA (ACE) images \cite{coburn2025acetada}, totaling 2,448 meals. Nutrition5k supplies fine-grained ingredient/component metadata, not a one-to-one visible-food item ontology \cite{thames2021nutrition5k}; a recent manual validation also documented omissions in its original labels
\cite{sterling2026vlmbenchmark}. Because our inventory unit is an
independently localizable and portionable visible food item, we denote its
recovery results N5k$^{*}$ and use the fixed 12,491 component references
only as a normalization denominator, not as item-level ground truth.
ACE uses 4,589 structured food-item references directly. Thus the
$\rho$ metrics are reference-normalized evidence-state rates instead of identity recall. The ``plate only'' sentinel is excluded from item and
numerical evaluation, consistent with prior Nutrition5k curation
\cite{bianco2024nutrition5k}. 

We evaluate two representative open-weight MLLMs from distinct families, Gemma~4 26B A4B (G26B) \cite{gemma4} and Qwen3.5-35B-A3B (Q35B) \cite{qwen35}, under the same FNDDS 2021--2023 grounding pipeline \cite{usda2024fndds}. Dataset-agnostic prompts, verification contracts, and SAM thresholds are frozen before formal evaluation and shared across datasets and models, without tuning on evaluation annotations. Visual grounding receives no reference labels or item counts, FNDDS codes, or nutrient
values.

For food-code grounding, BGE-based dense retrieval (BAAI/bge-small-en-v1.5 \cite{xiao2024bge}) and BM25 rankings are combined using weighted reciprocal-rank fusion \cite{cormack2009rrf}: $\mathrm{RRF}(c)=\sum_{r\in\mathcal{R}} w_r/(k+\operatorname{rank}_r(c))$, with $k=60$ and equal dense/sparse family weights. Retrieval depth is 50 per branch.

\subsection{Item-recovery evaluation}
\label{sec:itemrecoveryeval}

Recovery is evaluated on two evidence states, independently for each
model--dataset pair. \textbf{Initial} is the state after initial
proposal verification and before residual review. \textbf{Post-recovery} combines the initial verified evidence with source-blind verified recovery evidence, before deterministic reconciliation. Existing initial inventory identities remain represented regardless of their support state; a novel residual target is added only when recovery yields a successfully verified novel-item assignment. The same semantic and grouped-geometry audits from Sec.~\ref{sec:itemaudit} are then recomputed on the combined evidence.

Let $N_{\rm ref}$ denote the fixed dataset normalization count,
$N_{\rm rep}$ the number of represented item identities,
$N_s$ the number with $P_s=1$, and $N_g$ the number with $P_g=1$.
We report
\begin{equation}
\rho_{\rm item}=\frac{N_{\rm rep}}{N_{\rm ref}},\qquad
\rho_{\rm sem}=\frac{N_s}{N_{\rm ref}},\qquad
\rho_{\rm geo}=\frac{N_g}{N_{\rm ref}}.
\label{eq:recoverymetrics}
\end{equation}
For N5k$^{*}$, $N_{\rm ref}=12{,}491$; for ACE,
$N_{\rm ref}=4{,}589$. These are reference-normalized representation and verifier-assessed support rates, not ground-truth identity or segmentation scores.


\begin{table*}[!t]
\centering
\caption{Pre-reconciliation recovery on N5k$^{*}$/ACE: reference-normalized rates (\%); parentheses show changes from Initial (pp).}
\label{tab:recovery}
\small
\setlength{\tabcolsep}{4.0pt}
\renewcommand{\arraystretch}{0.85}

\begin{tabular}{@{}llccc@{}}
\toprule
Model & Evidence state &
$\rho_{\rm item}$ N5k$^{*}$/ACE &
$\rho_{\rm sem}$ N5k$^{*}$/ACE &
$\rho_{\rm geo}$ N5k$^{*}$/ACE \\
\midrule


\multirow{2}{*}{G26B}
& Initial
& 35.38/93.01
& 27.08/77.86
& 26.50/76.14 \\
& Post-recovery (ours)
& 35.44/93.07~(+0.06/+0.07~pp)
& 31.41/86.01~(+4.33/+8.15~pp)
& 30.85/84.24~(+4.36/+8.11~pp) \\

\midrule

\multirow{2}{*}{Q35B}
& Initial
& 35.80/90.98
& 27.70/80.67
& 27.52/80.58 \\
& Post-recovery (ours)
& 35.94/91.00~(+0.14/+0.02~pp)
& 32.77/87.60~(+5.07/+6.93~pp)
& 32.45/87.49~(+4.93/+6.91~pp) \\

\bottomrule

\end{tabular}
\end{table*}

\noindent\textbf{Manual food-identity audit:}
Separately from pre-reconciliation recovery evaluation, we manually
audit a fixed sample of 150 randomly selected, reference-screened
ACETADA C1 meals per model, from G26B/Q35B C1 cohorts of 472/680 meals.
Because N5k$^{*}$ component references do not map one-to-one to
visible-food items (Sec.~\ref{sec:implementation}), identity precision
and recall are reported only on ACE. The initial inventory and
Final-$F$ are evaluated on the same meals against the same reference
foods under strict one-to-one semantic matching. We report
micro-averaged precision and recall,

$P_{\rm micro}=\frac{\sum_n TP_n}{\sum_n(TP_n+FP_n)}$ and
$R_{\rm micro}=\frac{\sum_n TP_n}{\sum_n(TP_n+FN_n)}$.

\subsection{Nutrition baselines and matched evaluation}

For controlled end-to-end comparison, we use \textbf{Direct}, a whole-image nutrition estimate from the same MLLM without item-level grounding, and an \textbf{adapted DietAI24} reimplementation \cite{yan2025dietai24}. Following its meal-level retrieval design, adapted DietAI24 expands one holistic meal description into multiple queries and infers food codes from a shared FNDDS candidate pool; our method instead retrieves independently for each grounded Final-$F$ item. All three methods use the same evaluated MLLM; adapted DietAI24 and ours also share embeddings and the FNDDS release. Together, the two baselines provide controlled whole-image and retrieval-grounded comparisons; methods with different input, training, or 3D-resource assumptions are discussed qualitatively.

Because valid-output subsets differ across methods, the primary nutrition
comparison for model $m$ and dataset $d$ uses the common C1 intersection
\begin{equation}
S_{\cap}^{(m,d)}
=
S_{\rm Direct}^{(m,d)}
\cap
S_{\rm Adapted}^{(m,d)}
\cap
S_{\rm Ours,C1}^{(m,d)},
\label{eq:intersection}
\end{equation}
where $S_{\rm Ours,C1}^{(m,d)}$ contains complete-output C1 meals and the baseline sets contain valid outputs. Direct returns valid outputs for all meals, so adapted DietAI24 determines the intersection with our complete-output C1 set. C1+C2 provides broader-cohort context. We report meal-level mass and energy mean absolute error (MAE),

$\mathrm{MAE}_q=N^{-1}\sum_{j=1}^{N}
|\hat Y_{j,q}-Y_{j,q}|$,
for $q\in\{\mathrm{mass},\mathrm{energy}\}$.

Fig.~\ref{fig:scatter} uses Q35B/ACE C1 predictions without
restriction to the baseline intersection.
$\mathcal{R}_{\rm full}$ contains meals with
$r_i^{\rm sem}=r_i^{\rm geo}=1$ for every $i\in F_{\rm eff}$;
$\mathcal{R}_{\rm gap}$ contains the remainder.
\section{Results and Discussion}

\subsection{Item-level recovery}

Table~\ref{tab:recovery} compares the initial and post-recovery pre-reconciliation evidence states for G26B and Q35B. Across N5k$^{*}$ and ACE, recovery increases $\rho_{\rm item}$ by $0.02$--$0.14$ percentage points (pp), $\rho_{\rm sem}$ by $4.33$--$8.15$~pp, and $\rho_{\rm geo}$ by $4.36$--$8.11$~pp. These gains mainly reflect repaired semantic and geometric support for represented items, with limited additions of verified omitted foods.

Because N5k$^{*}$ uses component-level references, its absolute rates are not directly comparable with ACE; within-dataset changes are the relevant recovery signal.

\subsection{Final-\texorpdfstring{$F$}{F} identity quality}

Table~\ref{tab:human_audit} reports a manual audit of 150 ACETADA C1 meals per model using strict one-to-one semantic matching. Final-$F$ raises micro P/R from $98.35/93.02\%$ to $98.62/94.34\%$ for G26B and from
$90.54/87.19\%$ to $93.73/87.87\%$ for Q35B, improving both precision and recall. G26B begins near the precision ceiling ($98.35\%$). This audit evaluates the complete inventory-to-$F$ transformation, whereas
Table~\ref{tab:recovery} measures pre-reconciliation evidence support
instead of identity correctness.

\begin{table}[!t]
\centering
\caption{Manual food-identity audit on 150 ACETADA C1 meals per model.
Values are micro precision/recall (\%); superscripts show changes from Initial (pp).}
\label{tab:human_audit}
\small
\renewcommand{\arraystretch}{0.96}

\begin{tabular}{@{}l@{\hspace{6pt}}
c@{\hspace{7pt}}
c@{\hspace{11pt}}
c@{\hspace{7pt}}
c@{}}
\toprule
& \multicolumn{2}{c}{G26B} & \multicolumn{2}{c}{Q35B} \\
\cmidrule(lr){2-3}\cmidrule(lr){4-5}
Representation & $P_{\rm micro}$ & $R_{\rm micro}$
               & $P_{\rm micro}$ & $R_{\rm micro}$ \\
\midrule
Initial inventory
& 98.35
& 93.02
& 90.54
& 87.19 \\

Final-$F$ (Ours)
& \textbf{98.62}{\scriptsize$^{\uparrow0.27}$}
& \textbf{94.34}{\scriptsize$^{\uparrow1.32}$}
& \textbf{93.73}{\scriptsize$^{\uparrow3.19}$}
& \textbf{87.87}{\scriptsize$^{\uparrow0.68}$} \\
\bottomrule
\end{tabular}
\end{table}

\subsection{Matched-sample nutrition accuracy}

Table~\ref{tab:nutrition} compares Direct, adapted DietAI24, and our method on matched C1 cohorts. Ours achieves the lowest mass MAE in all four model--dataset settings and lower energy MAE than adapted DietAI24 in all four, while the energy comparison with Direct is dataset-dependent. These are end-to-end results, not isolated recovery effects. C1 denotes visual completeness but does not guarantee nutrition accuracy.

Across both datasets, the post-recovery C1 pools contain 1,351/2,448 meals ($55.2\%$) for G26B and 1,875/2,448 ($76.6\%$) for Q35B; the corresponding
C1+C2 pools contain 2,262 ($92.4\%$) and 2,220 ($90.7\%$), respectively. The matched C1 intersections contain $n_{\cap}=719$ (G26B) / $904$ (Q35B)
on N5k and $383$ (G26B) / $494$ (Q35B) on ACE. ACE C1+C2 meals route roughly twice as many effective Final-$F$ items as N5k ($5.1$ vs.\ $2.5$ per meal), consistent with the larger adapted DietAI24 gap on ACE, where a shared meal-level retrieval context must disambiguate more foods.

\begin{figure}[!t]
\centering
\includegraphics[width=\columnwidth]
{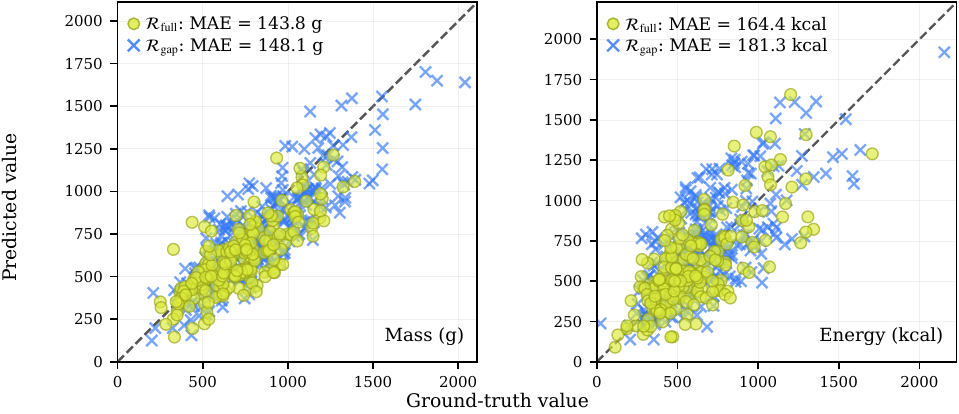}
\vspace{-5.8mm}
\caption{Q35B/ACETADA C1 mass and energy predictions without baseline-intersection filtering, grouped by $\mathcal{R}_{\rm full}$ and $\mathcal{R}_{\rm gap}$.}
\label{fig:scatter}
\end{figure}

Within Q35B/ACE C1 meals, Fig.~\ref{fig:scatter} shows lower MAE for the fully semantic- and geometry-ready group $\mathcal{R}_{\rm full}$ than for $\mathcal{R}_{\rm gap}$ in both mass ($143.8$ vs.\ $148.1$~g) and energy ($164.4$ vs.\ $181.3$~kcal).

\begin{table}[!t]
\centering
\caption{Meal-level MAE (mass g / energy kcal; lower is better). Direct, Adapted DietAI24, and Ours (C1) use the same matched C1 samples; Ours (C1+C2) reports our broader-cohort result.}
\label{tab:nutrition}
\small
\setlength{\tabcolsep}{2.2pt}
\renewcommand{\arraystretch}{0.80}

\begin{tabular}{@{}lccc@{\hspace{3pt}}c@{}}
\toprule
Model & Direct & Adapted DietAI24 & Ours (C1) & Ours (C1+C2) \\
\midrule

\multicolumn{5}{c}{\textit{Nutrition5k (N5k)}} \\
G26B & 81.1/119.3 & 123.7/164.9 &
\textbf{74.3}/\textbf{107.4} & 80.8/111.2 \\
Q35B & 114.3/138.8 & 150.8/189.1 &
\textbf{74.2}/\textbf{96.9} & 77.1/97.1 \\

\midrule

\multicolumn{5}{c}{\textit{ACETADA (ACE)}} \\
G26B & 217.8/\textbf{168.1} & 368.7/250.6 &
\textbf{140.5}/177.1 & 159.4/180.4 \\
Q35B & 192.5/\textbf{166.2} & 438.4/262.9 &
\textbf{144.8}/172.5 & 148.1/173.8 \\

\bottomrule
\end{tabular}
\end{table}

\subsection{Limitations and future work}

Visual completeness does not eliminate ambiguity in low-contrast
adjacent foods or mixed dishes. Verification adds inference overhead.
Motivated by the item-density and readiness patterns, future work
will quantify this cost and explore adaptive recovery for unresolved
items while preserving bounded inference for typical meals.
\section{Conclusion}
We presented a food-item verification and recovery framework for image-based
nutrition estimation without task-specific fine-tuning. Recovery primarily strengthens semantic and geometric support while changing represented-item
rates by at most $0.14$ percentage points. On matched C1 cohorts, our method achieves the lowest mass MAE across all four model--dataset settings and lower energy MAE than adapted DietAI24 in all four, while energy gains over Direct are dataset-dependent. These results reinforce the distinction between visual evidence quality and downstream nutrition accuracy.

\clearpage



\section{Compliance with Ethical Standards}
This study uses existing Nutrition5k and ACETADA data
\cite{thames2021nutrition5k,coburn2025acetada}
for computational evaluation. No new human or animal
experiments were conducted.

\bibliographystyle{IEEEbib}
\bibliography{references/references_compact}
\end{document}